\documentclass[fleqn,10pt]{wlscirep}
\usepackage[utf8]{inputenc}
\usepackage[T1]{fontenc}
\title{EHCTNet: Enhanced Hybrid of CNN and Transformer Network for Remote Sensing Image Change Detection}

\author[1]{Junjie Yang}
\author[1]{Haibo Wan}
\author[1,*]{Zhihai Shang}
\affil[1]{Lingnan Normal University, School of Geographical Sciences, Zhanjiang, 524048, China}
\affil[*]{shangzh@lingnan.edu.cn}

\begin{abstract}
Remote sensing (RS) change detection incurs a high cost because of false negatives, which are more costly than false positives. Existing frameworks, struggling to improve the Precision metric to reduce the cost of false positive, still have limitations in focusing on the change of interest, which leads to missed detections and discontinuity issues. This work tackles these issues by enhancing feature learning capabilities and integrating the frequency components of feature information, with a strategy to incrementally boost the Recall value. We propose an enhanced hybrid of CNN and Transformer network (EHCTNet) for effectively mining the change information of interest. Firstly, a dual branch feature extraction module is used to extract the multi-scale features of RS images. Secondly, the frequency component of these features is exploited by a refined module I. Thirdly, an enhanced token mining module based on the Kolmogorov-Arnold Network is utilized to derive semantic information. Finally, the semantic change information’s frequency component, beneficial for final detection, is mined from the refined module II. Extensive experiments validate the effectiveness of EHCTNet in comprehending complex changes of interest. The visualization outcomes show that EHCTNet detects more intact and continuous changed areas and perceives more accurate neighboring distinction than state-of-the-art models.
\end{abstract}
\begin{document}

\flushbottom
\maketitle
\thispagestyle{empty}

% \noindent Please note: Abbreviations should be introduced at the first mention in the main text – no abbreviations lists. Suggested structure of main text (not enforced) is provided below.

\section*{Introduction}

Remote sensing (RS) imagery, compared to traditional surveying and mapping data, offers unparalleled advantages such as wide-area coverage, high temporal frequency, fewer constraints on data collection, and rich information content. These characteristics make RS imagery well-suited for studying land use and environmental changes at large scales. Depending on the dataset characteristics, change detection tasks can be categorized into various types, including synthetic aperture radar (SAR), multi-spectral, hyper-spectral, very high-resolution, and heterogeneous image change detection~\cite{ShafiqueAyesha2022review}. Among these, very high-resolution imagery has gained prominence due to its ability to capture intricate details of artificial and natural structures. Consequently, methods for change detection in high-resolution RS images are critical for applications such as agricultural surveys, disaster assessment, land cover monitoring, urban expansion studies, and internal urban change analysis.

Despite its significance, change detection in RS imagery remains a challenging problem due to several factors: 1) limited spectral information in very high-resolution images compared to medium-resolution ones~\cite{DallaMura2015limitedspectralinformation}, 2) high spectral variability due to complex artificial materials~\cite{LU2016spectralvariability, ChenHao2022}, 3) information loss caused by occlusions such as clouds and shadows~\cite{TATAR2018informationloss}, and 4) inconsistent object features across images from different platforms, times, or locations~\cite{MercierGrÉgoire2008featureRepresentation, ChenHao2022, JiangBo2023VcT}. These challenges underscore the need for advanced models that can extract high-level semantic change information from bi-temporal images, surpassing traditional feature extraction approaches.

Traditional change detection methods relied heavily on manually crafted features and conventional machine learning techniques~\cite{RICHARDS1984PCA, Bovolo2008SVM, LiuSicong2015ChangeVectorAnalysis}, which often fail to capture complex spatial and temporal relationships in RS imagery~\cite{Cheng2024RSCDReview}. Deep learning-based approaches, particularly convolutional neural networks (CNNs) and Transformers, have since emerged as the dominant paradigms, achieving superior performance by automatically learning hierarchical features~\cite{miao2024dasunet, tang2024siamese}. While CNNs excel at extracting localized features, they struggle to capture global context due to their limited receptive fields~\cite{chen2022mobileformerbridgingmobilenettransformer}. In contrast, Transformers offer a global perception capability but often require large datasets to perform optimally, making them less suitable for small-scale datasets~\cite{Khan_2023SurveyofHVT}. 

To bridge these gaps, hybrid architectures combining CNNs and Transformers have shown great promise in remote sensing change detection~\cite{ChenHao2022, shen2023advancing}. These hybrid methods leverage the local feature extraction capabilities of CNNs and the global context modeling of Transformers to achieve robust representations. However, existing methods often rely on architectures designed for general computer vision tasks, which may not be optimized for the unique characteristics of RS imagery~\cite{Peng2023conformer, Zheng2023LFormer}. Additionally, many current models focus on both changed regions and consistent background information. While consistent background modeling reduces false positives, it risks overshadowing the primary goal of RS change detection: identifying changes critical for applications like disaster response and illegal construction monitoring. These scenarios demand high recall rates to ensure that all significant changes are detected, even at the cost of reduced precision.

In this work, we propose Enhanced Hybrid CNN-Transformer Network (EHCTNet), a novel architecture designed to address the aforementioned challenges by integrating local and global feature extraction with advanced semantic token learning. Our method incorporates spectral and spatial attention mechanisms, hybridizing CNNs and Transformers for superior performance in remote sensing change detection tasks. The main contributions of our work are as follows:
\begin{itemize}
    \item [\textcolor{black}{$\bullet$}] We propose a dual-branch hybrid architecture combining CNN and Transformer blocks to effectively integrate local and global features. This approach enhances multi-scale feature representation and significantly improves the accuracy of remote sensing change detection. 
    \item [\textcolor{black}{$\bullet$}] We design novel modules including a head residual fast Fourier transform (HFFT),  a KAN-based channel and spatial attention block (CKSA), and a back residual fast Fourier transform (BFFT), which are tailored to sequentially extract first-order features, semantic tokens, and second-order semantic difference information. These modules enhance the model's ability to capture both subtle and high-level changes, thereby improving the Recall metric and ensuring comprehensive detection.
    \item [\textcolor{black}{$\bullet$}] EHCTNet demonstrates robust performance in detecting changes of interest, maintaining the continuity and internal integrity of changed objects while enhancing discriminability between adjacent regions. Visualization results further highlight its effectiveness in remote sensing change detection tasks.
\end{itemize}

\begin{figure*}[t]
    \centering
    \includegraphics[width=1\linewidth]{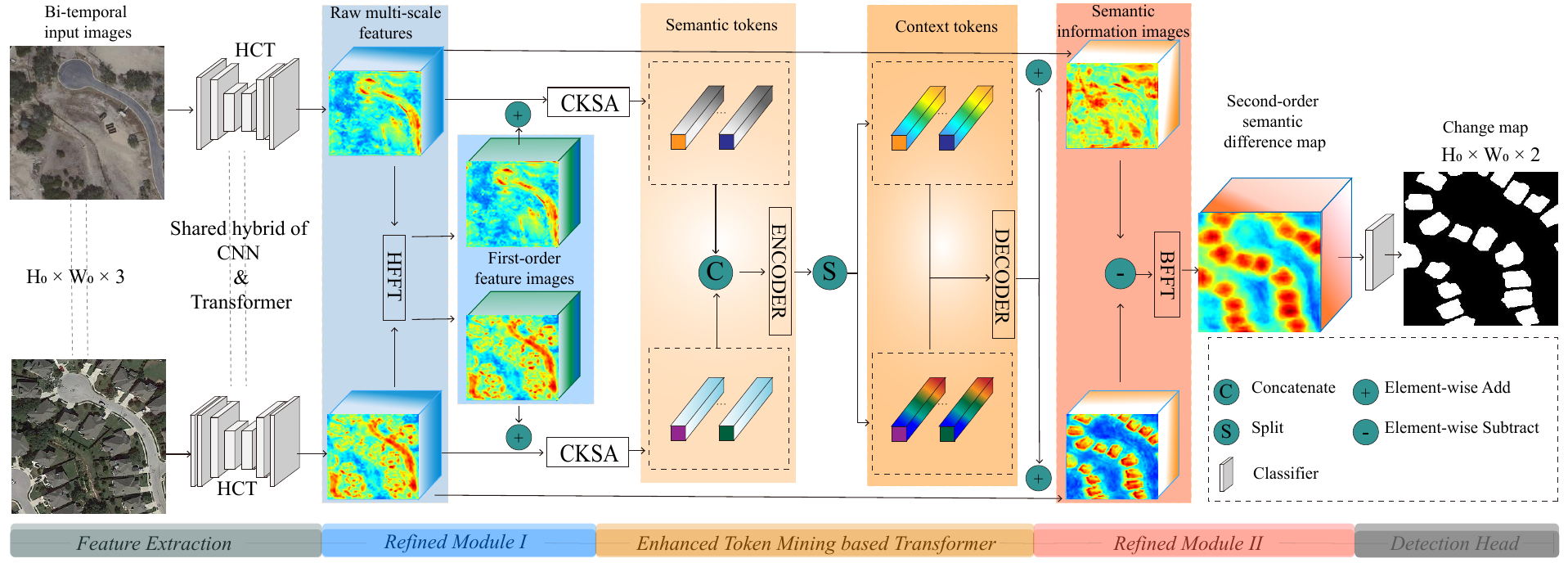}
\caption{Overall network structure of the EHCTNet.}
%\vspace{-.2cm}
\label{fig:model}
\end{figure*}

\section*{Related Work}\label{sec:rw}

\subsection*{Change Detection}
Remote sensing change detection methods can be broadly classified into two phases: conventional and deep learning-based approaches. Traditional statistical and machine learning techniques, such as image algebra~\cite{Bruzzone2000imageAlgebra}, principal component analysis (PCA)~\cite{Celik2009PCA, RICHARDS1984PCA}, clustering~\cite{LiuJunfu2020clustering}, change vector analysis~\cite{LiuSicong2015ChangeVectorAnalysis}, support vector machines (SVM)~\cite{Bovolo2008SVM}, random forests~\cite{Seo2018RandomForest}, and decision trees~\cite{Jungho2005DecisionTree}, dominate the conventional category~\cite{JiangBo2023VcT, Chen2024Cross-DomainReview}. These methods rely on manually designed features, offering interpretability but lacking the capacity to represent complex, high-level object information~\cite{Cheng2024RSCDReview}. Consequently, their detection capabilities are limited and are generally considered inferior to deep learning approaches~\cite{Chen2024Cross-DomainReview}.

Deep learning methods, such as convolutional neural networks (CNNs) and Transformers, have significantly advanced change detection by automatically learning hierarchical features~\cite{Cheng2024RSCDReview}. These methods can be categorized into single-stream, siamese, and generative adversarial networks (GANs)~\cite{Chen2024Cross-DomainReview}.
Single-stream models represent image pairs in a common feature space and utilize a single architecture for change detection. For example, Zhan et al. proposed a logarithmic transformation framework for heterogeneous SAR and optical images, achieving comparable statistical distributions between the modalities~\cite{Zhan2018SingStreamHeterogeneous}. Similarly, sparse autoencoders and capsule neural networks have been utilized to map images into shared feature spaces for improved classification~\cite{Ma2019SingStreamHeterogeneous, GONG2017SingStreamConditions}. While single-stream models are computationally efficient, they often fail to fully capture discriminative semantic information, limiting their effectiveness.
Siamese networks, on the other hand, utilize two sub-networks to extract features from bi-temporal images, followed by a change detection module. These networks, such as DSMNN-Net~\cite{Seydi2021RemoteSensing} and FC-Siam family architectures~\cite{Daudt2018IEEE, DAUDT2019CVIU}, excel at handling heterogeneous datasets and imaging conditions. Zhao et al. introduced a feature space transformation to align heterogeneous images for change detection~\cite{Zhao2017SiameseHeterogeneous}. However, the complexity of these architectures and the large number of parameters can lead to optimization challenges~\cite{Chen2024Cross-DomainReview}.
GAN-based approaches have recently gained attention for their ability to handle domain adaptation and generate synthetic data. For instance, CycleGANs have been employed to impose cycle-consistency for domain adaptation in SAR-optical change detection~\cite{LIU2022GANCondition}. Despite their potential, GANs suffer from instability during training, which can impact feature representation and model performance~\cite{Chen2024Cross-DomainReview}.

Although siamese and GAN-based methods show promise, their complexity and training challenges highlight the need for more efficient and robust solutions.

\subsection*{Hybridization Architectures}
The hybridization of CNNs and Transformers has emerged as a powerful approach for feature learning in computer vision~\cite{Khan_2023SurveyofHVT}. While pre-trained CNNs are often used as backbones for feature extraction in change detection~\cite{ChenHao2022, Li2022TransUNet, JiangBo2023VcT}, their performance can be limited when applied directly to remote sensing tasks. To address this, hybrid architectures integrate CNNs for local feature extraction with Transformers for capturing global relationships~\cite{Peng2023conformer, Zheng2023LFormer}.

For example, Zheng et al. proposed L-Former, which employs Transformers in shallow layers and CNNs in deeper layers, achieving competitive results~\cite{Zheng2023LFormer}. However, existing hybrid models often overlook the importance of capturing consistent background information to reduce false positives in change detection tasks~\cite{JiangBo2023VcT}. Recent works have emphasized focusing on background consistency to improve precision and F1 scores~\cite{Jian2022GANCondition, JiangBo2023VcT}. Recently, diffusion models~\cite{shen2024imagdressing,shen2024imagpose} have demonstrated exceptional capabilities in generating fine-grained, semantically consistent outputs across various domains, including computer vision and natural language processing. Yet, this shift risks neglecting the primary goal of identifying changed areas, particularly in critical applications such as disaster response and unauthorized construction monitoring, where recall is paramount.

Despite the progress in hybrid models, existing methods lack sophisticated strategies tailored to specific remote sensing scenarios. To bridge this gap, we propose an enhanced hybrid CNN-Transformer network that improves feature learning capability and overall accuracy. By incorporating token mining and refinement modules, our method incrementally enhances recall, ensuring comprehensive and reliable change detection.

\section*{Proposed Method}\label{sec:method} 
The entire architecture of EHCTNet, comprising five modules: 1) the feature extraction module, 2) refined module I, 3) the enhanced token mining based Transformer module, 4) refined module II, and 5) the detection head module, which is depicted in Fig.\ref{fig:model}. The feature extraction module consists of dual branches hybridization architecture of CNN and Transformer (HCT) designed to capture raw multi-scale features from bi-temporal images. The HCT combines the local feature extraction capabilities of CNNs with Transformers' global contextual feature learning abilities to significantly enhance the raw feature representation. Refined module I, which follows the feature extraction module, is a frequency attention module designed to refine the frequency details of the raw multi-level features and to generate first-order features. The enhanced token mining based Transformer module adopts the first-order refined features as input to gain semantic information. Refined module II, which is placed in the deeper part of EHCTNet, symmetrically to refined module I, is also a frequency attention module designed to refine the frequency components of the semantic information within the enhanced token mining based Transformer module and to generate second-order semantic difference information. Refined module I primarily aids the model in acquiring refined frequency features for each image, which benefits change detection, and refined module II is utilized for learning high-level semantic difference information from the semantic difference map. Lastly, the detection head is used to generate the change map.

\subsection*{Feature extraction module} \label{subsec:feature extraction module} 
\begin{figure}[t]
    \centering
    \begin{minipage}{0.45\linewidth}
        \centering
        \includegraphics[width=\linewidth]{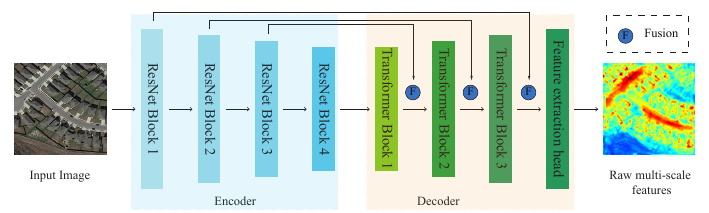}
        \caption{Illustration of the HCT branch.}
        \label{fig:feature extraction module}
    \end{minipage}
    \hfill
    \begin{minipage}{0.45\linewidth}
        \centering
        \includegraphics[width=\linewidth]{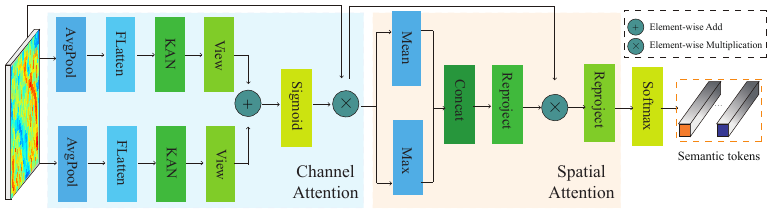}
        \caption{Illustration of CKSA.}
        \label{fig:CKSA}
    \end{minipage}
\end{figure}

Inspired by the hybridization idea~\cite{Khan_2023SurveyofHVT} and the CMTFNet model~\cite{Wu2023HVTRSSemantic}, we construct a dual HCT blocks, named the feature extraction module, to fuse local features with global features. The two branches of the feature extraction module, which share learnable parameters, are identical. Each branch of the feature extraction module is an HCT, and the structure of the HCT is shown in Fig. \ref{fig:feature extraction module}. The encoder part uses ResNet50 to encode hierarchical features, while the decoder part consists of 3 Transformer blocks to decode multi-scale global contextual features. In order to fuse the local hierarchical features with the global contextual features, fusion operations are performed after the first three decoder blocks. The first two fusion operations capture rich local features and global contextual information but lack spatial detail. Therefore, the third fusion is crucial for integrating the spatial features from the first CNN module, such as radiation intensities, edges, corners, and texture. The decoder generates multi-scale global contextual information and gradually restores the feature resolution by fusing the hierarchical features obtained from the CNN module. In addition, a learnable variable is used to weigh the importance of the local features and global contextual information in the fusion process. Thus, the contribution of the two elements to the output is formulated as:
\begin{equation}
    O = \alpha \times O_E + (1-\alpha)\times O_D,
\end{equation}
where $O$ denotes the output of fusion, $\alpha$ represents the learnable variable, $O_E$ is the output local features of the encoder part, and $O_D$ is the output global contextual information of the decoder part.

\subsection*{Refined module I}\label{subsec:refined module I} 
Combining spectral layers and multi-head attention enables models to achieve state-of-the-art (SOTA) performance~\cite{patro2023efficiency, patro2023spectformer}. Therefore, we designed joint modules consisting of refined module I, the enhanced token mining based Transformer module, and refined module II, which follow the feature extraction module. Refined module I yields first-order features that are beneficial for representing detailed information in each raw feature image. In this work, the refined module I mainly consists of a fast Fourier transform (FFT) layer, a weighted gating mechanism, and an inverse FFT layer~\cite{patro2023spectformer}, which can be expressed as:
\begin{equation}
\label{equation: FFT}
X_{FFT} = IFFT(Gate(FFT(X_{FE}))),
\end{equation}
where $X_{FFT}$ is the output of the $FFT$, $Gate$, and $IFFT$ operations, $X_{FE}$ is the raw feature image, $FFT$, $Gate$, and $IFFT$ represent the FFT, the weighted gating mechanism and the inverse FFT procedure within refined module I.

The FFT layer in Equation~\ref{equation: FFT} transforms the feature map from the physical to the spectral space. The weighted gating mechanism, as a learnable weight parameter within the neural network, can effectively identify the frequency-domain features in the feature map by adjusting its weights through the process of back-propagation during training, thereby determining the significance of each frequency component in the feature representation. The inverse FFT converts the feature map from the spectral back to the spatial domain, thereby generating refined frequency features with enhanced detail, referred to as first-order features. 

Finally, the output of refined module I is passed through a residual connection to retain the characteristics in the raw feature image. The process mentioned above can be expressed as follows:
\begin{equation}
X_{HFFT} = X_{FFT} + X_{FE},
\end{equation}
where $X_{HFFT}$ is the output of the refined module I.

\subsection*{Enhanced token mining based Transformer module}\label{subsec:enhanced token mining based Transformer module}
The feature extraction module extracts and fuses multi-scale features from bi-temporal RS images in the previous two procedures. Then, we utilize refined module I to obtain the first-order features. In this step, the enhanced token mining based Transformer module is used for the semantic token extraction and semantic information perception. The enhanced token mining based Transformer module consists of two units: the KAN~\cite{liu2024kan} based channel and spatial attention (CKSA) block, and the Transformer unit. 

Semantic token operations are beneficial for interacting with change information in remote sensing change detection tasks~\cite{ChenHao2022,JiangBo2023VcT}. Semantic tokens represent high-level concepts of change interest. It is one of the key elements in change detection. In addition, the applicability and efficacy of KAN have already been tested in computer vision domain~\cite{azam2024suitability} and RS domain~\cite{cambrin2024kan}. Drawing inspiration from the KAN layer’s ability to facilitate the learning of tailored activations at the network edge and compute the contribution of each input channel~\cite{liu2024kan,cambrin2024kan}, we first designed the CKSA block (Fig. \ref{fig:CKSA}) to generate two condense token sets for precise learning of semantic tokens within this module. The CKSA block mainly comprises a channel and spatial attention units. Specifically, we replace the fully connected layer with a KAN layer. In the CKSA, the process of converting feature images to tokens can be expressed as:
\begin{equation}
X_{CKSA} = SA(CKA(X_{HFFT})),
\end{equation}
where $X_{CKSA}$ is the output token of CKSA, CKA means the KAN-based channel attention operation, and SA means the spatial attention operation.

CKSA obtains tokens of image features, which contain rich details of changes in the feature images but lack interactive relational semantic information between the tokens. Transformer~\cite{vaswani2017attention} can fully exploit high-level global semantic relations in the token space~\cite{ChenHao2022}. Therefore, we incorporate a Transformer block into the subsequent stage of the enhanced token mining based Transformer module. First, the two token sets derived from the CKSA are concatenated to form a token aggregation, which is then fed into the encoder of the Transformer~\cite{ChenHao2022, JiangBo2023VcT} to capture the global semantic context between these tokens. Since the token aggregation involves concatenating semantic token sets along the second dimension (dim = 1), it can be likened to binding two bundles of token bars together. Therefore, the Transformer encoder can extract the intra-relationships within one set of tokens and the inter-relationships between the two sets of semantic tokens. As a result, the output from the Transformer encoder is enriched with high-level semantic information within the intra-tokens and global semantic information across the inter-tokens.

We divide the high-level semantic context into two sets of context tokens, each with the same dimension as the CKSA tokens. The two sets of context tokens encapsulate condensed semantic context and represent the high-level information regarding the hotspot. Subsequently, Transformer decoder~\cite{vaswani2017attention} was deployed to restore the two sets of context tokens to dual semantic pixel maps in pixel space. The dual-branch pixel maps, rich in high-quality semantic information, enable each pixel within the maps to be represented by the two sets of context tokens. This representation effectively highlights the pixel values of interest in the semantic maps.

The semantic pixel maps effectively reveal semantic hotspots in the feature space. Subsequently, the semantic pixel maps are subtracted from each other to obtain a semantic difference map, which represents the semantic information of change. Subtraction between the semantic pixel maps can result in both positive and negative outcomes. To ensure that all values are non-negative, we convert any negative results to their absolute values, as defined by:
\begin{equation}
X_{SUB} = |X_{SPFM1}-X_{SPFM2}|,
\end{equation}
where $X_{SPFM1}$ and $X_{SPFM2}$ represent the pixel values of the semantic pixel maps, and $X_{SUB}$ denotes the pixel value of the semantic difference map generated by taking the absolute of the subtraction.

\begin{table}[t]
\centering
\begin{minipage}{0.45\linewidth}
    \caption{Comparisons with other SOTA models on LEVIR-CD datasets.  The best and second results are marked in \textcolor{red}{RED} and \textcolor{blue}{BLUE}, respectively. All these scores are written in percentage (\%).}
    \label{benchmarkResultsLEVIR-CD}
    \begin{tabular}{lccccc}
    \toprule
    \textbf{Method} & Rec. & F1 & IoU & OA  \\
    \midrule
    \textbf{FC-Siam-Conc}~\cite{daudt2018fully} & 76.77 & 83.69 & 71.96 & 98.49 \\
    \textbf{VcT}~\cite{JiangBo2023VcT} & 77.45 & 85.04 & 86.26 & 98.61 \\   
    \textbf{FC-EF}~\cite{daudt2018fully}        & 80.17 & 83.40 & 71.53 & 98.39 \\
    \textbf{IFNet}~\cite{zhang2020deeply}        & 82.93 & 88.13 & 78.77 & \textcolor{blue}{\bf98.87}  \\
    \textbf{FC-Siam-Di}~\cite{daudt2018fully}   & 83.31 & 86.31 & 75.92 & 98.67 \\
    \textbf{BIT}~\cite{ChenHao2022} & 84.84 & 88.40 & \textcolor{blue}{\bf89.01} & \textcolor{blue}{\bf98.87} \\   
    \textbf{DTCDSCN}~\cite{liu2020building}      & 86.83 & 87.67 & 78.05 & 98.77 \\    
    \textbf{SNUNet}~\cite{fang2021snunet}       & 87.17 & 88.16 & 78.83 & 98.82 \\
    \textbf{CropLand}~\cite{liu2022cnn}      & \textcolor{blue}{\bf87.57} & \textcolor{blue}{\bf88.67} & 79.64 & 98.86 \\    
    % \hline   
    \multicolumn{5}{c}{\dotfill} \\
    \textbf {Baseline}    &82.95   &87.15     &87.96     &98.75\\
    \textbf {EHCTNet (Ours)} & \textcolor{red}{\bf88.53} & \textcolor{red}{\bf90.00} & \textcolor{red}{\bf90.38} & \textcolor{red}{\bf99.00} \\
    \bottomrule
    \end{tabular}
\end{minipage}
\hfill
\begin{minipage}{0.45\linewidth}
    \caption{Comparisons with other SOTA models on DSIFN-CD datasets.}
    \label{benchmarkResultsDSIFN-CD}
    \begin{tabular}{lcccc}
    \toprule
    \textbf{Method} & Rec. & F1 & IoU & OA \\
    \midrule
    \textbf{FC-EF}~\cite{daudt2018fully}   & 52.73 & 61.09 & 43.98 & 88.59\\
    \textbf{IFNet}~\cite{zhang2020deeply}  & 53.94 & 60.10 & 42.96 & 87.83 \\
    \textbf{FC-Siam-Conc}~\cite{daudt2018fully} & 54.21 & 59.71 & 42.56 & 87.57\\
    \textbf{VcT}~\cite{JiangBo2023VcT} & 63.52 & 72.50 & \textcolor{blue}{\bf73.85} & \textcolor{blue}{\bf91.81} \\ 
    \textbf{CropLand}~\cite{liu2022cnn} & 65.08 & 60.53 & 43.40 & 87.03\\   
    \textbf{FC-Siam-Di}~\cite{daudt2018fully}  & 65.71 & 62.54 & 45.50 & 86.63\\
    \textbf{SNUNet}~\cite{fang2021snunet} & 72.89 & 66.18 & 49.45 & 87.34\\
    \textbf{BIT}~\cite{ChenHao2022}  & 73.18 & \textcolor{blue}{\bf73.23} & 73.69 & 90.91\\
    \textbf{DTCDSCN}~\cite{liu2020building} & \textcolor{blue}{\bf77.99} & 63.72 & 46.76 & 84.91\\    
    % \hline    
    \multicolumn{5}{c}{\dotfill} \\
    \textbf {Baseline}   &70.82   &74.49     &74.99    &91.76\\
    \textbf {EHCTNet (Ours)} & \textcolor{red}{\bf89.01} & \textcolor{red}{\bf79.30} & \textcolor{red}{\bf78.20} & \textcolor{red}{\bf92.10} \\
    \bottomrule
    \end{tabular}
\end{minipage}
\end{table}

% 尝试用我自己的图
\begin{figure*}[t]
    \centering
    \includegraphics[width=1\linewidth]{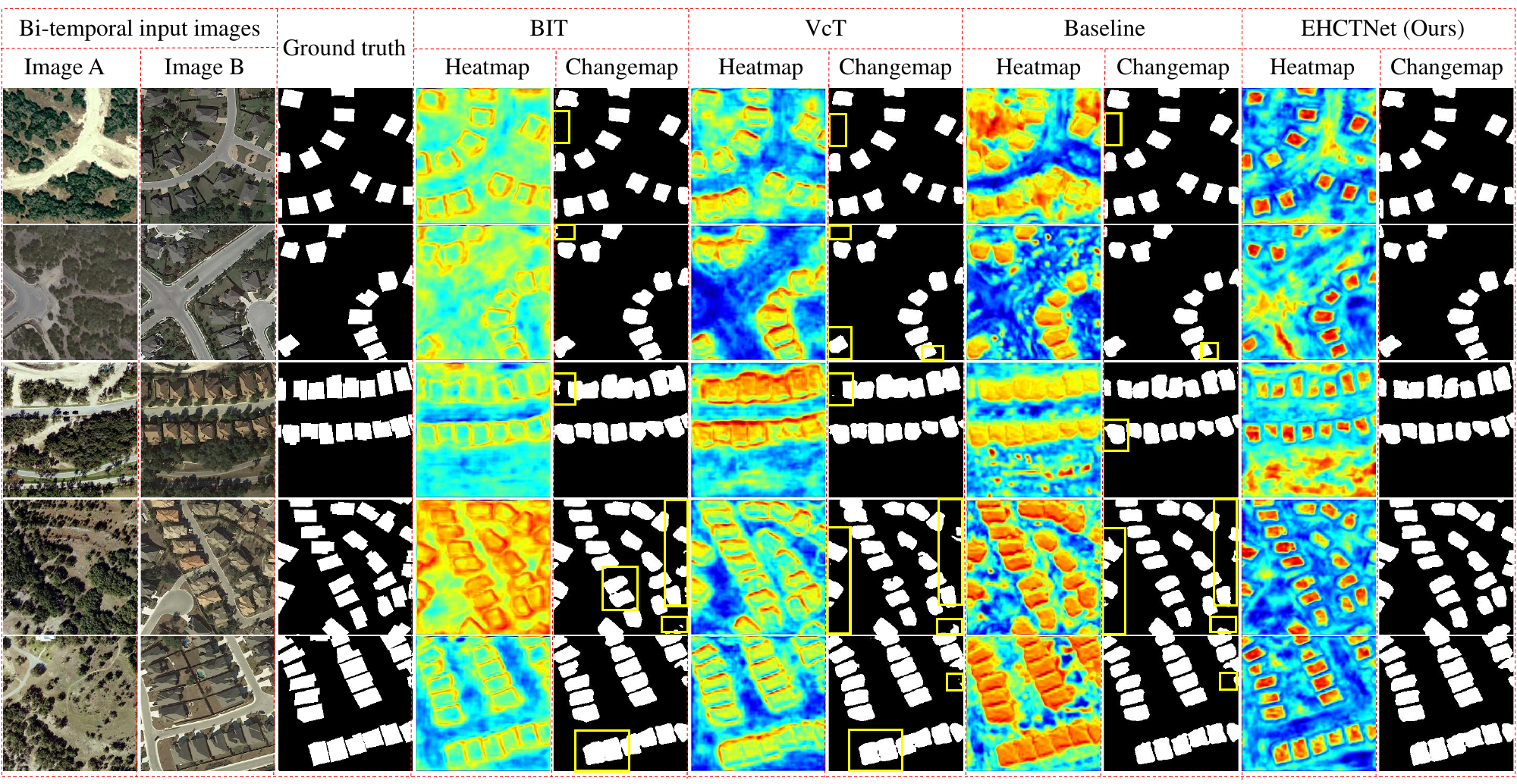}
\caption{The heatmaps and change results of our EHCTNet are compared with those of two state-of-the-art (SOTA) approaches (BIT~\cite{ChenHao2022} $\&$ VcT~\cite{JiangBo2023VcT}). The baseline means the feature extraction module of our EHCTNet. The yellow rectangle indicates the presence of missed detections, false detections, and incorrect merging of adjacent change areas.}
% The heatmaps were generated by the average of all bands of results of each method.
%\vspace{-.2cm}
\label{fig: Heatmap Comparing with SOTA}
\end{figure*}

\subsection*{Refined module II}\label{subsec:refined module II}
HFFT of refined module I and BFFT of refined module II are symmetrically positioned at the early and late stages of the EHCTNet model, as illustrated in Figure~\ref{fig:model}. Similar to HFFT, BFFT also includes an FFT layer, a weighted gating, an inverse FFT layer, and a residual connection. BFFT is used to refine the semantic difference map and generate the second-order semantic difference map. The refined module II can be expressed as follows:
\begin{equation}
X_{BFFT} = IFFT(Gate(FFT(X_{SUB}))) + X_{SUB},
\end{equation}
where $X_{BFFT}$ is the output of the refined module II.
% The difference between the refined module I and refined module II lies in the use of the HFFT. In refined module I, the HFFT is shared by the dual branches of feature images extracted from the feature extraction module. In contrast, in refined module II, the BFFT receives only the semantic difference map, which is the result of the subtraction, as its input.

In this subsection, the semantic difference map is first rescaled to match the dimensions of the original RS images. The BFFT converts the rescaled semantic difference map's physical space to spectral space, where it portrays the detailed information of the semantic difference map. It then restores this depicted detail information to the physical space, thereby generating second-order semantic difference information in the refined semantic difference map. 

\subsection*{Detection head}\label{subsec:detection head Module}
The second-order semantic difference information in the refined semantic difference map represents the final stage of semantic information. It is directly utilized in the detection head module to discriminate between the changed and background areas. A full convolutional network is employed in the detection head to generate a change map, which has a dimension of $\mathbb{R}^{H_{0} \times W_{0} \times 2}$, where $H_{0}$ and $W_{0}$ represent the height and width of the original bi-temporal RS images.

\subsection*{Loss function}  
Currently, most change detection tasks are still binary classification problems. The cross-entropy loss function is one of the most favored methods for minimizing the loss in neural networks and optimizing the network parameters. By definition, the loss function is formally expressed as:
\begin{equation}
L(G,P)=-\frac{1}{H_{0}\times W_{0}}\sum_{i=1}^{H_{0}\times W_{0}}G(i)\log P(i),
\end{equation}
where $L$ is the cross-entropy loss, $G$ represents the ground truth value and $P$ denotes the predicted value. %The $H_0$ and $W_0$ denote the height and width of input images respectively.

\section*{Experiment and Analysis}\label{sec:exp} 
To comprehensively validate the superiority of our proposed EHCTNet model, we have conducted extensive comparisons with SOTA change detection methods on two large-scale, high spatial resolution remote sensing (RS) image datasets, namely, \textbf{\emph{LEVIR-CD}}~\cite{ChenHao2022} and \textbf{\emph{DSIFN-CD}}~\cite{zhang2020deeply}.

\subsection*{Datasets}
\textbf{\emph{LEVIR-CD}} is a large RS dataset for building change detection. The original dataset derived from Google Earth contains 637 pairs of very high-resolution RS images, each with a size of 1024 × 1024 pixels. The images in the LEVIR-CD dataset were cut into small patches of size 256 × 256 due to the limitations of GPU power~\cite{ChenHao2022}, resulting in 7120, 1024, and 2048 pairs for training, validation, and testing, respectively, in this study.
%\textbf{\emph{WHU-CD}}

\textbf{\emph{DSIFN-CD}} is a publicly available binary change detection dataset that comprises high-resolution (2m) satellite image pairs from six major Chinese cities: Beijing, Chengdu, Shenzhen, Chongqing, Wuhan, and Xi’an. These image pairs are manually curated from Google Earth and are split into 394 sub-image pairs of 512×512 pixels. Subsequently, these sub-image pairs are augmented to yield 3988 dual-temporal image pairs. The dataset is divided into 3600 training pairs, 340 validation pairs, and 48 testing pairs. This dataset is extensively utilized in the RS change detection domain. Similar to the LEVIR-CD dataset, these 512 × 512 DSIFN-CD image pairs were further sliced into 256 × 256 patches to accommodate the GPU memory constraints during the work, resulting in subsets of 14400 training image pairs, 1360 validation image pairs, and 192 testing image pairs for change detection~\cite{JiangBo2023VcT}.

\begin{figure*}[t]
    \centering
    \begin{minipage}{0.45\linewidth}
        \centering
        \includegraphics[width=\linewidth]{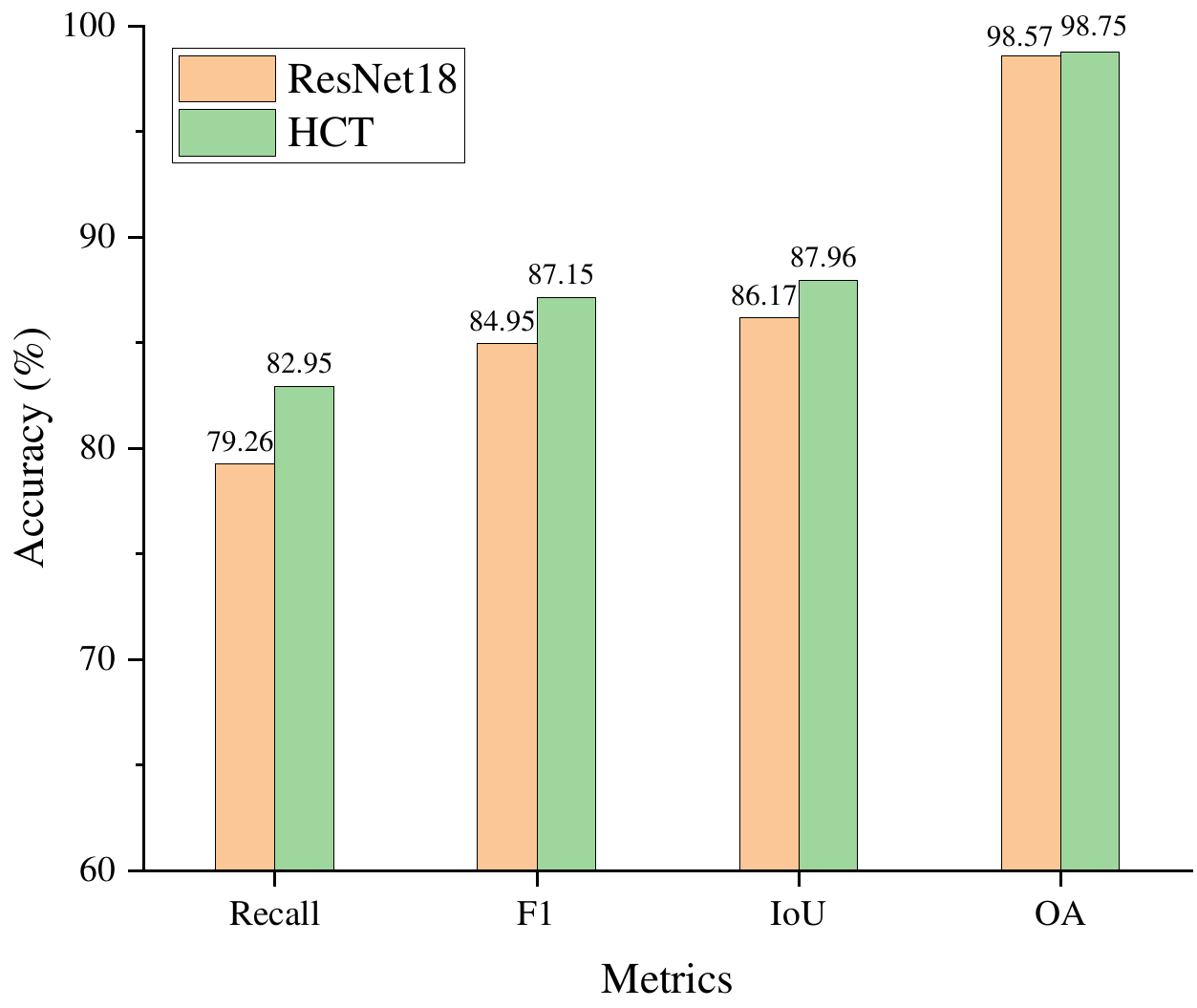}
        \caption{The improvement of our HCT on LEVIR-CD dataset over ResNet18.}
        \label{fig:Comparison Bar Chart of Feature Extraction Modules on the LEVIR-CD Dataset}
    \end{minipage}
    \hfill
    \begin{minipage}{0.45\linewidth}
        \centering
        \includegraphics[width=\linewidth]{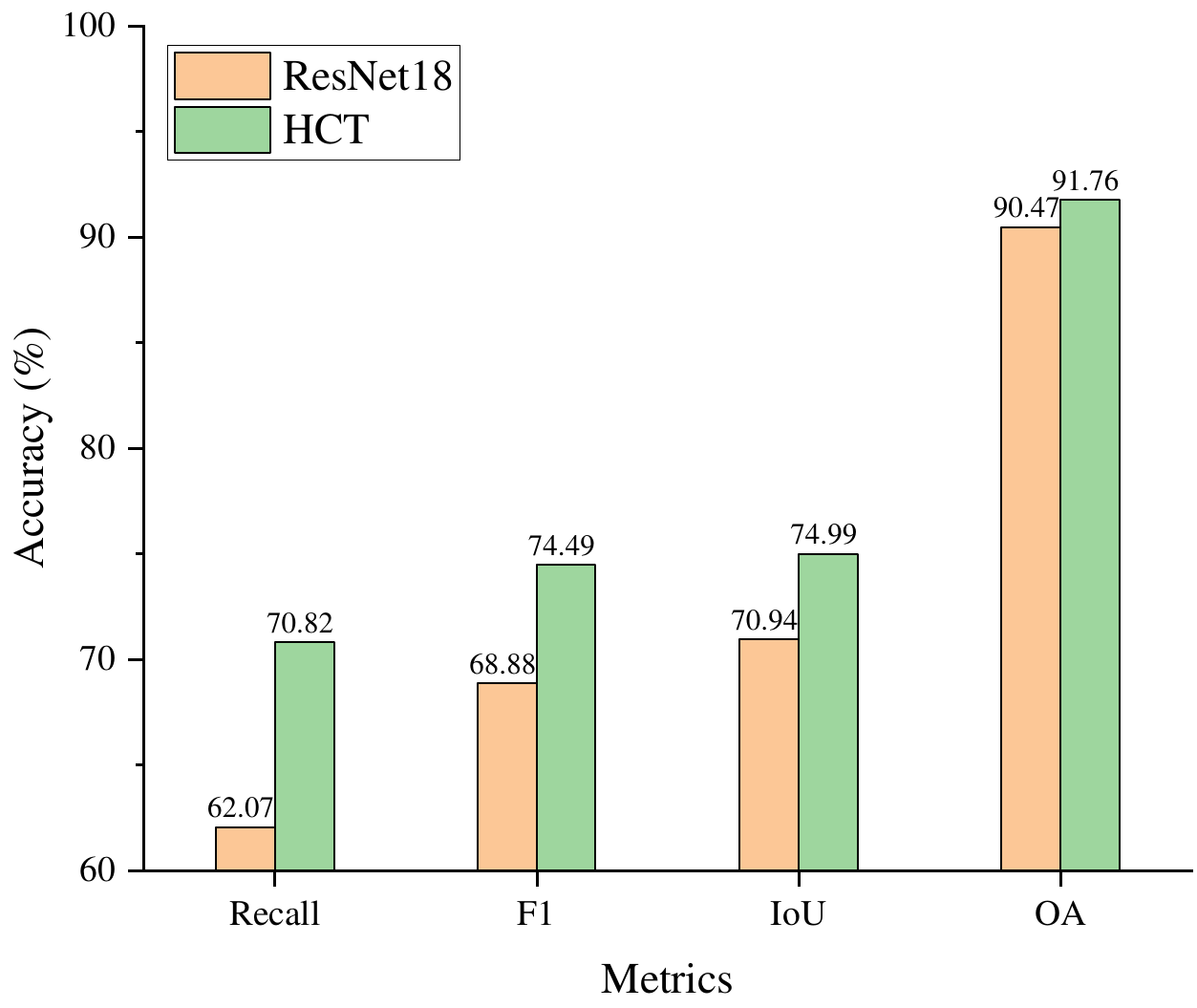}
        \caption{The improvement of our HCT on DSIFN-CD dataset over ResNet18.}
        \label{fig:Comparison Bar Chart of Feature Extraction Modules on the DSIFN-CD Dataset}
    \end{minipage}
\end{figure*}

\subsection*{Evaluation metrics}  
The standard evaluation indices used in the change detection tasks to assess the performance of approaches include \textbf{Precision}, \textbf{Recall}, \textbf{Intersection over Union} (IoU), and \textbf{Overall Accuracy} (OA)~\cite{ChenHao2022}.

\textbf{Precision} quantifies the proportion of correct positive predictions, defining it as the percentage of correctly predicted positive instances out of all the instances predicted as positive, which is defined as:
\begin{equation}
Precision = TP/(TP + FP),
\end{equation}
where $TP$ and $FP$ represent the number of true and false positives in the confusion matrix computed by comparing the ground truth with the predicted values. When the cost or risk associated with false positives is high (High Cost of False Positives), it is crucial to maximize Precision as much as possible. This is particularly important in medical diagnosis, legal decisions, and financial fraud detection. In this case, a high Precision indicates that the algorithm is less likely to misclassify background information as areas of change.

\textbf{Recall} represents the number of positive predictions made out of all actual positive cases in the dataset, and it quantifies the percentage of actual positives that were correctly identified, which is defined as:
\begin{equation}
Recall = TP/(TP+FN),
\end{equation}
where $FN$ represents the false negative value in the confusion matrix. In contrast to Precision, Recall is more appropriate for cases where the cost of false negatives is high, such as in disease screening, security identification, rare events detection, and change detection. In this case, a high Recall indicates that the model successfully identifies most of the change of interest. In change detection scenarios, it is crucial to achieve a high Recall to ensure that the change of interest is not overlooked, as the occurrence of change is significant and rare.

\textbf{IoU} is commonly used to measure the proportion of overlap between the ground truth and the predicted results, which is expressed as:
\begin{equation}
\label{equation: IoU}
IoU = TP/(TP+FN+FP).
\end{equation}

The IoU value ranges from 0 to 1, with 1 indicating perfect overlap between the ground truth and the prediction and 0 indicating no overlap. In change detection tasks, we prioritize the changed area over the background. As stated in Formula~\ref{equation: IoU}, the IoU score primarily focuses on the area of interest that has changed rather than considering both categories (changed area and background area) equally. Furthermore, after calculating the IoU scores for all regions of interest, it is necessary to compute the average of these scores to obtain the mean Intersection over Union (\textbf{MIoU}), which is defined as:
\begin{equation}
MIoU = \frac{1}{N} \sum_{i=1}^{N} {IoU}_i,
\end{equation}
where $i$ denotes the index of the region of interest, $N$ is the total number of regions of interest.

\textbf{OA}, which stands for Overall Accuracy, represents the percentage of correctly predicted areas out of the total areas in the dataset. It is defined as:
\begin{equation}
OA = (TP+TN)/(TP+TN+FN+FP),
\end{equation}
where $TN$ represents true negative. OA calculates an algorithm's performance across all classes into a single value; however, it does not provide insight into the algorithm's effectiveness on individual classes, especially in the case of imbalanced datasets.
%下午对Precision、Recall、IoU和OA进行详细解释。2024.10.1

Historically, the \textbf{F1-score} has been considered a primary evaluation metric in remote sensing change detection tasks due to its consideration of both Precision and Recall. This is reflected in works such as Chen et al. (2022)~\cite{ChenHao2022} and Jiang et al. (2023)~\cite{JiangBo2023VcT}, where the F1-score is formulated as follows:
\begin{equation}
F1 = 2 * \frac {Precision * Recall} {Precision + Recall}.
\end{equation}
% However, to evaluate the effectiveness of our proposed model in the specific domain of change detection, we not only consider the F1-score but also focus on Recall and IoU to detect areas of interest for change as comprehensively as possible.

In conclusion, the Recall, F1-score, and IoU metrics are suitable for evaluating the performance of models in change detection tasks. The OA metric is also adopted in this work to understand the overall performance of change detection models across both change and background categories.

\subsection*{Implementation details} 
EHCTNet is implemented in PyTorch and trained end-to-end on an NVIDIA GeForce RTX 3060 GPU using the stochastic gradient descent (SGD) optimizer and a linear learning rate policy. The initial learning rate is 0.01, the batch size is 8, the weight decay is 0.0005, and the momentum is 0.99, respectively. After each training epoch, a validation process is conducted, and the model with the best performance is saved for evaluation on the test set.

\begin{table}[t]
\centering
\begin{minipage}{0.45\linewidth}
    \caption{Ablation study of our EHCTNet on LEVIR-CD dataset. Ablations are experimented on our baseline, enhanced token mining based Transformer (ETMT), refined module I (RMI) and refined module II (RMII).}
    \label{DifferentComponentsAnalysisLEVIR-CD}
    \begin{tabular}{lccccc}
    \toprule
    \textbf{Method} & Rec. & F1 & IoU & OA  \\
    \midrule
    \textbf {Baseline}    &82.95   &87.15     &87.96     &98.75\\
    % \textbf {FE+TMT}    &85.26   &88.99   &89.52   &98.93\\
    \textbf {+ETMT}     &87.56   &89.51    &89.96    &98.96\\
    \textbf {+RMI+ETMT}     &88.22   &89.58    &90.02     &98.96\\
    \textbf {+ETMT+RMII}     &\textcolor{blue}{\bf88.41}   &\textcolor{blue}{\bf89.71}    &\textcolor{blue}{\bf90.13}    &\textcolor{blue}{\bf98.97}\\
    \textbf {EHCTNet}     &\textcolor{red}{\bf88.53}   &\textcolor{red}{\bf90.00}    &\textcolor{red}{\bf90.38}     &\textcolor{red}{\bf99.00}\\
    \bottomrule
    \end{tabular}
\end{minipage}
\hfill
\begin{minipage}{0.45\linewidth}
    \caption{Ablation study of our EHCTNet on DSIFN-CD dataset. Ablations are experimented on our baseline, enhanced token mining based Transformer (ETMT), refined module I (RMI) and refined module II (RMII).}
    \label{DifferentComponentsAnalysisDSIFN-CD}
    \begin{tabular}{lcccc}
    \toprule
    \textbf{Method} & Rec. & F1 & IoU & OA \\
    \midrule
    \textbf {Baseline}   &70.82   &74.49     &74.99    &91.76\\
    % \textbf {FE+TMT}   &87.34   &78.91   &77.92   &92.07\\
    \textbf {+ETMT}    &87.75   &79.28   &\textcolor{blue}{\bf78.26}    &\textcolor{blue}{\bf92.21}\\
    \textbf {+RMI+ETMT}   &88.03  &\textcolor{red}{\bf79.59}    &\textcolor{red}{\bf78.53}    &\textcolor{red}{\bf92.33}\\
    \textbf {+ETMT+RMII}    &\textcolor{blue}{\bf88.74}   &77.62    &76.59   &91.30\\
    \textbf {EHCTNet}   &\textcolor{red}{\bf89.01}  &\textcolor{blue}{\bf79.30}    &78.20   &92.10\\
    \bottomrule
    \end{tabular}
\end{minipage}
\end{table}

\subsection*{Comparison with State-of-the-art methods} 
We compare EHCTNet with 9 SOTA methods in this subsection on the LEVIR-CD and DSIFN-CD benchmark datasets. These SOTA methods include FC-EF~\cite{daudt2018fully}, FC-Siam-Di~\cite{daudt2018fully}, FC-Siam-Conc~\cite{daudt2018fully}, DTCDSCN~\cite{liu2020building}, IFNet~\cite{zhang2020deeply}, SNUNet~\cite{fang2021snunet}, CropLand~\cite{liu2022cnn},  BIT~\cite{ChenHao2022}, VcT~\cite{JiangBo2023VcT}. The first four methods are CNN-based architectures, while the rest are Transformer-based. In our study, we selected BIT and VcT as the primary comparison benchmarks due to their distinct contributions to the field of change detection. BIT stands out as a cutting-edge, robust Transformer architecture specifically designed for change detection tasks. Concurrently, VcT is distinguished by its focus on leveraging background information, which has recently shown superior performance compared to other methods in the domain~\cite{JiangBo2023VcT}.
%These methods are implemented based on public code with default hyperparameters~\cite{ChenHao2022, JiangBo2023VcT}. The optimal parameters mentioned in their work ~\cite{ChenHao2022, JiangBo2023VcT} are used in BIT and VcT, to generate results on LEVIR-CD and DSIFN-CD datasets respectively.%

\subsubsection*{Comparisons on LEVIR-CD}
The result of the above nine change detection methods and our EHCTNet on the LEVIR-CD dataset is listed in Table~\ref{benchmarkResultsLEVIR-CD}. Table~\ref{benchmarkResultsLEVIR-CD} shows that the EHCTNet architecture significantly outperforms the other approaches. The Recall, F1, IoU, and OA metrics of EHCTNet achieve the best scores among the above nine methods. In specific, our EHCTNet outperforms BIT by +3.69$\%$, +1.60$\%$, +1.37$\%$, +0.13$\%$ on Recall, F1, IoU, and OA, and beats VcT by +11.38$\%$, +4.96$\%$, +4.12$\%$, +0.39$\%$, respectively. These improved metrics demonstrate the effectiveness of the EHCTNet framework for RS image change detection tasks. Furthermore, the Recall metrics indicate that the EHCTNet framework can identify changed areas to the extent practicable, rendering it highly effective for specific RS image change detection tasks, particularly for datasets such as LEVIR-CD, which can be used as training data sets for the detection of illegal constructions.

\subsubsection*{Comparisons on DSIFN-CD}
The experimental results of the different methods on the DSIFN-CD dataset are shown in Table~\ref{benchmarkResultsDSIFN-CD}. Table~\ref{benchmarkResultsDSIFN-CD}  shows that our EHCTNet method achieved the highest values for Recall, F1, IoU, and OA metrics compared to the other nine approaches. Specifically, our EHCTNet outperformed BIT on Recall, F1, IoU, and OA by +15.83$\%$, +6.07$\%$, +4.51$\%$, and +1.19$\%$, respectively. In addition, the Recall, F1, IoU, and OA metrics of EHCTNet have been improved by +25.49$\%$, +6.80$\%$, +4.35$\%$, and +0.29$\%$ respectively, when compared with the VcT method. Especially for the Recall metric, the performance improvement indicates that EHCTNet achieves a significant improvement over VcT, which suggests that our EHCTNet does effectively detect essential and helpful change information from the dual images. Consequently, the EHCTNet method is suitable for changed area detection, such as detecting increments in artificial structures and facilities on the DSIFN-CD dataset.

Our EHCTNet mainly focuses on mining semantic change information from bi-temporal RS images. The comparison results in Table~\ref{benchmarkResultsLEVIR-CD}, and Table~\ref{benchmarkResultsDSIFN-CD} indicate that our EHCTNet not only excels these stable SOTA methods but also outperforms the VcT approach, which mainly pays close attention to the background information. It is worth noting that the Recall metric of EHCTNet substantially outperforms all the above models, illustrating that our EHCTNet is more suitable for the specific RS image change detection domain. It effectively captures the continuous and intact changes of interest, as demonstrated in Fig.~\ref{fig: Heatmap Comparing with SOTA}. 

\begin{figure*}[t]
    \centering
    \includegraphics[width=1\linewidth]{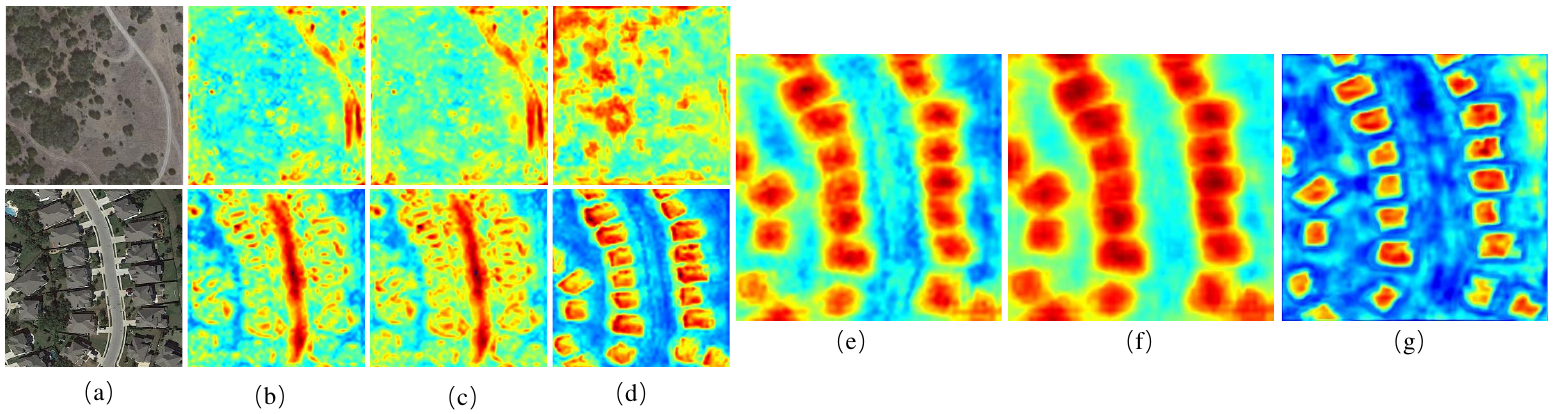}
\caption{An example of network visualization ((a): Bi-temporal input images; (b): Raw multi-scale feature images; (c): First-order feature images; (d): Semantic pixel maps; (e): Semantic difference map; (f): Second-order semantic difference map; (g): Change detection heatmap).}
%\vspace{-.2cm}
\label{fig:EHCTNet visualization}
\end{figure*}

\subsection*{Feature extraction improvement}
The HCT is utilized in the feature extraction module of our EHCTNet. We conducted a comparison experiment between the feature extraction module (ResNet 18) of SOTA methods and our HCT on the LEVIR-CD  dataset, as shown in Fig~\ref{fig:Comparison Bar Chart of Feature Extraction Modules on the LEVIR-CD Dataset}. It shows that our HCT significantly outperforms the ResNet commonly used in other SOTA change detection models. In Fig~\ref{fig:Comparison Bar Chart of Feature Extraction Modules on the LEVIR-CD Dataset}, our feature extraction module has improved Recall, F1, IoU, and OA by +3.69$\%$, +2.2$\%$, +1.79$\%$ and +0.18$\%$, respectively.

In addition, another comparison of DSIFN-CD datasets is also experimented with in this work, as shown in Fig~\ref{fig:Comparison Bar Chart of Feature Extraction Modules on the DSIFN-CD Dataset}. Fig~\ref{fig:Comparison Bar Chart of Feature Extraction Modules on the DSIFN-CD Dataset} shows that the HCT in EHCTNet is also superior to the ResNet 18. Compared to the ResNet 18, our HCT has improved Recall, F1, IoU, and OA by +8.75$\%$, +5.61$\%$, +4.05$\%$ and +1.29$\%$, respectively.

Note that the metrics of our hybrid feature extraction module improved significantly, which extracts more detailed image features than those feature extraction modules used in other change detection tasks. Our feature extraction baseline, which is only utilized for extracting image features in the shallow layers of the EHCTNet, even achieved competitive performance compared to other well-designed methods in Table~\ref{benchmarkResultsLEVIR-CD} and Table~\ref{benchmarkResultsDSIFN-CD}. The CNN-Transformer hybridization’s competitive performance ensures that the feature extraction module provides more detailed features to perceive high-level change information in subsequent procedures. Consequently, compared with other methods shown in Table~\ref{benchmarkResultsLEVIR-CD} and Table~\ref{benchmarkResultsDSIFN-CD}, EHCTNet achieves the highest evaluation metrics. Especially for the LEVIR-CD dataset, which accurately marks the edges of image objects, EHCTNet gained an extremely high IoU score of 90.38$\%$.

\subsection*{Ablation studies and analysis}  
\subsubsection*{Influence of enhanced token mining based Transformer module}

The feature extraction module extracts image features from dual temporal RS images using the multi-scale feature-extracting capabilities of the CNN and Transformer hybridization. This procedure comprehensively utilized dual images' spatial, textual, and brightness features. These image features are generally regarded as low-level features. Therefore, we integrate an enhanced token mining based Transformer module into EHCTNet to capture semantic information. The CKSA block is constructed to generate tokens by categorizing features into concepts, resulting in two bundles of tokens. The Transformer encoder exploits the global semantic relationships between tokens to generate semantic information about the area of interest. Subsequently, the Transformer decoder projects this high-level semantic information back into the feature image space to obtain pixel-level, semantic feature maps. The effectiveness of the enhanced token mining based Transformer module is shown in Table~\ref{DifferentComponentsAnalysisLEVIR-CD} and Table~\ref{DifferentComponentsAnalysisDSIFN-CD}. As can be seen from Table~\ref{DifferentComponentsAnalysisLEVIR-CD}, after introducing the enhanced token mining based Transformer module into EHCTNet, the Recall, F1 score, IoU, and OA of the network improved by +4.61$\%$, +2.36$\%$, +2.00$\%$, +0.21$\%$, respectively. Table~\ref{DifferentComponentsAnalysisDSIFN-CD} further shows that the Recall, F1, IoU, and OA of the network improved by +16.93$\%$, +4.79$\%$, +3.27$\%$, +0.45$\%$ respectively. These improvements demonstrate that the semantic information benefits change detection on the LEVIR and DSIFN datasets.
% % 改到这里了2024.11.15日晚10:00，明天从这里继续吧
% Therefore, the Transformer encoder is capable of extracting not only the intra-relationships within one set of tokens but also the inter-relationships between the two sets of semantic tokens. As a result, the output from the Transformer Encoder is enriched with high-level semantic information within the intra-tokens, as well as global semantic information across the inter-tokens.

\subsubsection*{Role of refined module I}
The feature extraction module can only provide raw image features to the enhanced token mining based Transformer module. In order to delve deeper into the frequency information behind the feature images, we develop the refined module I, placing it between the feature extraction module and the enhanced token mining based Transformer module. This module is designed to extract first-order features, such as lines, edges, corners, and other intricate patterns from feature images. By analyzing these localized frequencies, the network gains a more nuanced understanding of the extent and the transitional boundaries of each object feature within the image. Table~\ref{DifferentComponentsAnalysisLEVIR-CD} shows that the refined module I improves the Recall, F1 score, IoU, and OA by +0.66$\%$, +0.07$\%$, +0.06$\%$, +0.00$\%$ on the LEVIR-CD dataset, respectively. While Table~\ref{DifferentComponentsAnalysisDSIFN-CD} indicates that the refined module I improves the  Recall, F1 score, IoU, and OA by +0.28$\%$, +0.31$\%$, +0.27$\%$, +0.12$\%$ on the DSIFN-CD dataset, respectively. Therefore, these frequency components from the dual feature images provide an additional frequency variation feature to the subsequent modules of the EHCTNet model, which is conducive to the expression of change features. 

\subsubsection*{Impact of refined module II}
The enhanced token mining based Transformer module generates dual semantic pixel maps. An absolute subtraction of dual semantic maps is performed to generate a semantic-level difference map. Similar to refined module I, refined module II also employs an FFT layer, a weighted gating mechanism, and an inverse FFT layer to extract the frequency components of the semantic difference map. Table~\ref{DifferentComponentsAnalysisLEVIR-CD} shows that refined module II can improve the model’s performance (Recall: +0.85$\%$, F1 score: +0.20$\%$, IoU: +0.17$\%$, OA: +0.01$\%$) on the LEVIR-CD test dataset. Additionally, the combination of refined module I and refined module II (EHCTNet) achieves the best evaluation metrics, demonstrating the effectiveness of refined module I and refined module II on the LEVIR-CD test dataset. 

Although the DSIFN-CD test dataset is more complex than the LEVIR-CD dataset, refined module II can also perceive the valuable change information, significantly improving the Recall metric by +0.99$\%$. Furthermore, the combination of refined module I and refined module II also demonstrates the best performance on the DSIFN-CD dataset, achieving the highest Recall metric (89.01$\%$) under the premise of overall high evaluation metrics, as shown in Table~\ref{DifferentComponentsAnalysisDSIFN-CD}. 

The frequency component is added back to the semantic level difference map to enhance additional information that is beneficial for detecting change of interest. Table~\ref{DifferentComponentsAnalysisLEVIR-CD} and Table~\ref{DifferentComponentsAnalysisDSIFN-CD} show that our EHCTNet architecture utilized the ability of refined module I and refined module II can improve the Recall performance by 0.97$\%$ and 1.26$\%$ on LEVIR-CD and DSIFN-CD datasets respectively. Our EHCTNet is appropriate for improving the change detection rate in the specific remote sensing change detection tasks.

\subsection*{Visualization}  
% In the subsequent sections, the performance of each module of EHCTNet on the LEVIR-CD and DSIFN-CD datasets is comprehensively analyzed to investigate the logic behind its superiority. The effectiveness of the refined module I, the enhanced token mining based Transformer module, and refined module II of the EHCTNet model is tested by overlaying each module in turn. 
\begin{figure*}[t]
    \centering
    \includegraphics[width=1\linewidth]{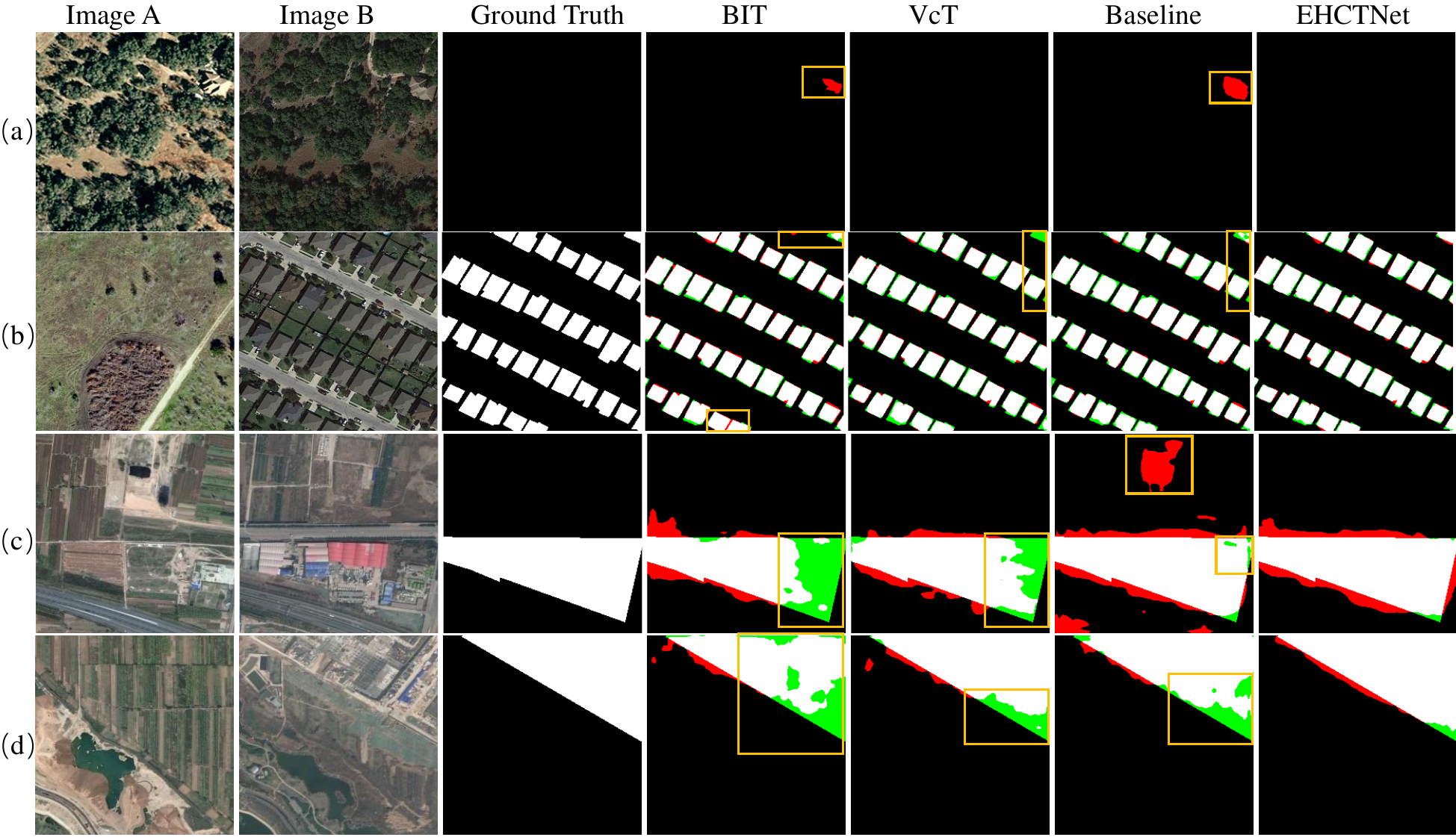}
\caption{Visualization results of different methods on the LEVIR-CD ((a) and (b)) and DSIFN-CD ((c) and (d)) test datasets. Different colors are utilized to highlight the performance differences among these methods in change detection tasks: white for true positives, black for true negatives, red for false positives, and green for false negatives. The orange rectangle indicates the presence of missed detections, discontinuity, and incorrect merging of adjacent change areas.}
%\vspace{-.2cm}
\label{fig:visualization results of different methods on the LEVIR-CD and DSIFN test dataset}
\end{figure*}

\subsubsection*{Network visualization}
In EHCTNet, several crucial nodes visualize the features and semantic information, including the raw features, first-order features, semantic pixel features, high-level semantic difference features, second-order semantic difference features, and change heatmap. These visualizations demonstrate the processing of dual-temporal images and help to understand the effectiveness of EHCTNet, as illustrated in Fig.~\ref{fig:EHCTNet visualization}. The feature extraction module of EHCTNet is used to extract raw features, which contain only low-level features. First-order features, perceived by the HFFT layer, enhance the frequency component of the raw features in the second step. Next, the enhanced token mining based Transformer is employed to learn semantic information of interest related to change. The semantic pixel maps can capture more continuous information of changed areas than the previous two steps, as shown in (d) of Fig.~\ref{fig:EHCTNet visualization}. Subsequently, the subtraction operation and BFFT are introduced to generate the high-level semantic difference map and the second-order semantic difference map, respectively. The change representations of the high-level semantic difference map and the second-order semantic difference map are more intact than the semantic pixel maps, as shown in (e) and (f) of Fig.~\ref{fig:EHCTNet visualization}. Finally, the change map is obtained by the detection head of EHCTNet, and a compact detection heatmap is shown in (g) of Fig.~\ref{fig:EHCTNet visualization}. This indicates that EHCTNet can progressively perceive features and semantic concepts in changed areas.

\subsubsection*{Visualization comparison to SOTA}
We compared four methods: two SOTA methods, our baseline, and our EHCTNet on the LEVIR-CD and DSIFN-CD test datasets, as shown in Fig.~\ref{fig:visualization results of different methods on the LEVIR-CD and DSIFN test dataset}. The two SOTA methods include BIT and Vct. BIT is typical of comprehensive SOTA methods that focus on the background and change information of dual temporal images. At the same time, VcT, as a SOTA approach, strives to mine the background information to detect changed areas. In Fig.~\ref{fig:visualization results of different methods on the LEVIR-CD and DSIFN test dataset}, rows (a) and (b) represent scenes of few buildings and dense buildings. Scene (a), which represents the two different light condition images of the same region, shows that both BIT and our baseline can detect the existence of buildings in the given context, but they cannot accurately perceive the unchanged information. While our EHCTNet can perceive that there is no change of interest in this scene. Rows (b) contains some cases of missed detection and lacks the ability to distinguish between neighboring objects, as indicated by the orange rectangle in row (b) for the BIT method. In addition, Our EHCTNet demonstrates the best performance in continuous and intact detection among the compared methods, as shown in (c) and (d) of Fig.~\ref{fig:visualization results of different methods on the LEVIR-CD and DSIFN test dataset}. Generally speaking, continuous and complete detection capability is highly advantageous in remote sensing change detection tasks, particularly in scenarios such as disaster area identification, victim rescue operations, illegal building monitoring, and asset loss assessment. In summary, our EHCTNet excels in prioritizing the detection of change of interest over BIT and VcT, and it can more clearly identify intact and continuously changed areas than either of the two methods. In summary, it provides more accurate neighboring distinction and edge information detection capability than SOTA methods.

\section*{Conclusion}\label{sec:con}
In this work, we proposed EHCTNet, an enhanced hybrid CNN-Transformer framework for remote sensing change detection. EHCTNet consists of five modules: feature extraction, refined modules I and II, an enhanced token mining-based Transformer, and a detection head. It combines CNNs and Transformers to extract multi-scale features, uses a KAN-based attention block to identify valuable tokens, and employs a symmetric FFT structure to refine frequency components of feature images and semantic maps. The detection head predicts changes between bi-temporal images.
Results showed that EHCTNet enhances edge detection, improves detection continuity, and achieves higher Recall, identifying more subtle changes than state-of-the-art methods. It effectively captures changes caused by complex reconstructions and solar radiation. Extensive experiments on LEVIR-CD and DSIFN-CD validated its effectiveness, with ablation studies highlighting the contributions of each module.
This work focused on single-source data, limiting its application to multi-modal scenarios. Future work will explore mining changes from multi-source remote sensing images to further enhance its versatility and effectiveness.

\section*{Data availability}
The LEVIR-CD and DSIFN-CD datasets are openly available at: \url{https://github.com/Event-AHU/VcT_Remote_Sensing_Change_Detection} (accessed on 29 January 2024).

\bibliography{sample}

% \noindent LaTeX formats citations and references automatically using the bibliography records in your .bib file, which you can edit via the project menu. Use the cite command for an inline citation, e.g.  \cite{Hao:gidmaps:2014}.

% For data citations of datasets uploaded to e.g. \emph{figshare}, please use the \verb|howpublished| option in the bib entry to specify the platform and the link, as in the \verb|Hao:gidmaps:2014| example in the sample bibliography file.

% \section*{Acknowledgements (not compulsory)}

% Acknowledgements should be brief, and should not include thanks to anonymous referees and editors, or effusive comments. Grant or contribution numbers may be acknowledged.

\section*{Author contributions statement}
Conceptualization, Z.S., H.W. and J.Y.; methodology, J.Y. and H.W.; validation, J.Y.; formal analysis, J.Y. and Z.S.; investigation, Z.S and J.Y.; resource, Z.S. and H.W.; data curation, J.Y.; writing-original draft preparation, J.Y. and H.W.; writing-review and editing, J.Y., H.W. and Z.S. All authors have read and agreed to the published version of the manuscript.

\section*{Funding}
The research was supported by the Scientific Research Project of Lingnan Normal University (ZL2401, ZL1935), the Ministry of Education of the People's Republic of China University-Industry Collaborative Education Program (220802313194815).

\section*{Competing interests }
The authors declare no competing interests.

\section*{Additional information}
\textbf{Correspondence} and requests for materials should be addressed to Z.S.

% To include, in this order: \textbf{Accession codes} (where applicable); \textbf{Competing interests} (mandatory statement). 

% The corresponding author is responsible for submitting a \href{http://www.nature.com/srep/policies/index.html#competing}{competing interests statement} on behalf of all authors of the paper. This statement must be included in the submitted article file.

% \begin{figure}[ht]
% \centering
% \includegraphics[width=\linewidth]{stream}
% \caption{Legend (350 words max). Example legend text.}
% \label{fig:stream}
% \end{figure}

% \begin{table}[ht]
% \centering
% \begin{tabular}{|l|l|l|}
% \hline
% Condition & n & p \\
% \hline
% A & 5 & 0.1 \\
% \hline
% B & 10 & 0.01 \\
% \hline
% \end{tabular}
% \caption{\label{tab:example}Legend (350 words max). Example legend text.}
% \end{table}

% Figures and tables can be referenced in LaTeX using the ref command, e.g. Figure \ref{fig:stream} and Table \ref{tab:example}.

\end{document}